\documentclass{article} 
\usepackage{iclr2026_conference,times}

\usepackage{amsmath,amsfonts,bm}

\def\eqref#1{equation~\ref{#1}}

\def\1{\bm{1}}

\DeclareMathAlphabet{\mathsfit}{\encodingdefault}{\sfdefault}{m}{sl}
\SetMathAlphabet{\mathsfit}{bold}{\encodingdefault}{\sfdefault}{bx}{n}

\usepackage{hyperref}
\usepackage{url}

\usepackage{amsmath,amssymb,amsfonts}
\usepackage{graphicx}
\usepackage{textcomp}
\usepackage{xcolor}
\usepackage[caption=false]{subfig}

\usepackage{booktabs}  
\usepackage{array}     

\usepackage{algorithm}
\usepackage{algpseudocode}
\algrenewcommand\alglinenumber[1]{\footnotesize #1:}
\algrenewcommand\algorithmicrequire{\textbf{Input:}}
\algrenewcommand\algorithmicensure{\textbf{Output:}}

\providecommand{\keywords}[1]{\par\noindent\textbf{Keywords:} #1}

\title{Boosting AI Reliability with an FSM-Driven Streaming Inference Pipeline: An Industrial Case\\
}

\author{Yutian Zhang$^{1*}$, Zhongyi Pei$^{1*\dagger}$, Yi~Mao$^{2}$, Chen Wang$^{1}$, Lin Liu$^{1}$, Jianmin Wang$^{1}$ \\
$^1$ School of Software, BNRist, Tsinghua University, China \\
$^2$ Tianyi Technology Co., Ltd, China \\
\texttt{yt-zhang24@mails.tsinghua.edu.cn} \\
\texttt{peizhyi@tsinghua.edu.cn} \\
\texttt{maoyi@tycmc.net} \\
\texttt{\{wang\_chen,linliu,jimwang\}@tsinghua.edu.cn} \\
}

\iclrfinalcopy 

\begin{document}

\maketitle

\begingroup
\renewcommand\thefootnote{}
\footnotetext{* denotes equal contribution. $\dag$ denotes the corresponding author.}
\addtocounter{footnote}{-1}
\endgroup

\begin{abstract}
The widespread adoption of AI in industry is often hampered by its limited robustness when faced with scenarios absent from training data, leading to prediction bias and vulnerabilities. To address this, we propose a novel streaming inference pipeline that enhances data-driven models by explicitly incorporating prior knowledge. 
This paper presents the work on an industrial AI application that automatically counts excavator workloads from surveillance videos. Our approach integrates an object detection model with a Finite State Machine (FSM), which encodes knowledge of operational scenarios to guide and correct the AI's predictions on streaming data. In experiments on a real-world dataset of over 7,000 images from 12 site videos, encompassing more than 300 excavator workloads, our method demonstrates superior performance and greater robustness compared to the original solution based on manual heuristic rules. We will release the code at \url{https://github.com/thulab/video-streamling-inference-pipeline}.
\end{abstract}


\section{Introduction}

In recent years, machine learning has achieved significant advancements across various domains \cite{canhoto2020artificial, app}. Despite their broad potential, machine learning models often struggle to meet complex business requirements in real-world settings. Two key challenges hinder their practical deployment. First, building end-to-end solutions for certain tasks, such as accurately interpreting logical information from video streams, remains difficult. Second, in dynamic environments, real-world data often diverges substantially from training data due to variations in background, lighting, and other contextual factors, leading to unreliable predictions and exacerbated model biases.

The reliability of machine learning systems has been studied in several research areas. Recent efforts in requirements engineering for AI (RE4AI) \cite{PeiLWW22}, solution patterns \cite{nalchigar2018business}, and MLOps \cite{mlops2022} have sought to systematize the development of robust AI applications. These works highlight the importance of non-functional requirements \cite{nfr_2021} and quality assessment models \cite{qa_2020,qa_2022}. Meanwhile, techniques such as data augmentation \cite{da}, adversarial training \cite{at}, and regularization \cite{reg} aim to enhance model robustness. However, these methods are primarily applied during training and often fail to address unforeseen scenarios encountered during real work.

\begin{figure}[!t]
\centering
\begin{minipage}[t]{0.23\linewidth}
  \centering
  \includegraphics[width=\linewidth,height=0.18\textheight,keepaspectratio]{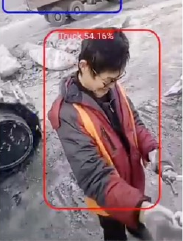}\\
  \vspace{2pt}
  (a) Incorrectly detects a person as a truck.
\end{minipage}\hfill
\begin{minipage}[t]{0.23\linewidth}
  \centering
  \includegraphics[width=\linewidth,height=0.18\textheight,keepaspectratio]{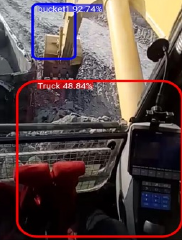}\\
  \vspace{2pt}
  (b) Incorrectly detects a pilothouse as a truck.
\end{minipage}\hfill
\begin{minipage}[t]{0.23\linewidth}
  \centering
  \includegraphics[width=\linewidth,height=0.18\textheight,keepaspectratio]{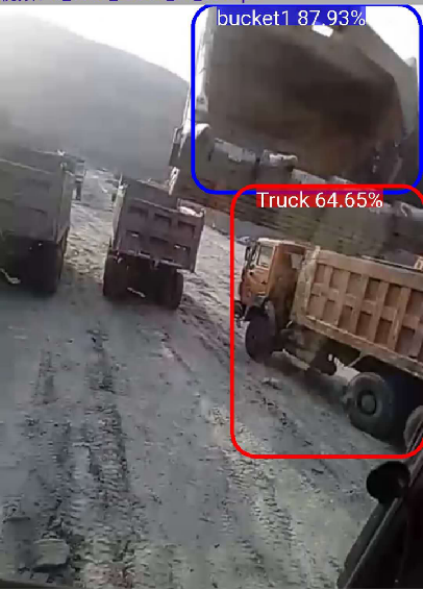}\\
  \vspace{2pt}
  (c) Incorrectly detects a non-target truck.
\end{minipage}\hfill
\begin{minipage}[t]{0.23\linewidth}
  \centering
  \includegraphics[width=\linewidth,height=0.18\textheight,keepaspectratio]{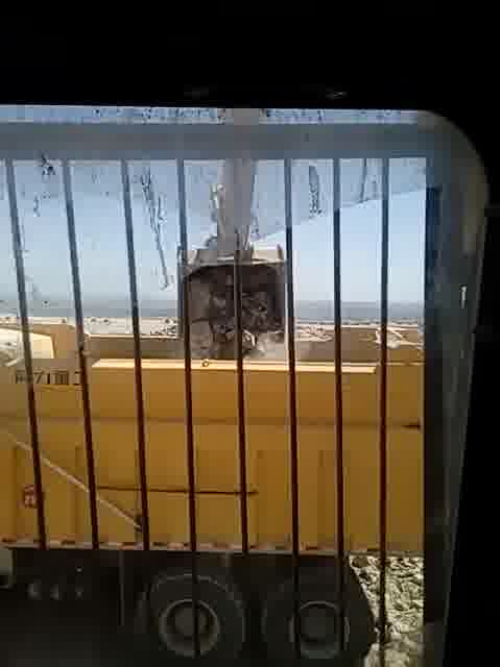}\\
  \vspace{2pt}
  (d) Truck detection under guardrail occlusion.
\end{minipage}
\caption{Wrong detection results lead to counting fake workloads, which often occur due to unseen objects, non-target vehicles, and occlusions not covered in the training set.}
\label{fig:badex}
\end{figure}

In this paper, we tackle the reliability challenge in a real industrial context, automatic excavator workload monitoring from surveillance videos. Excavators play an essential role in construction, and the leasing market demands accurate usage tracking to support rational pricing and improved utilization. Automatically counting excavator workloads, defined as a non-trivial task whose complete work cycle comprises (1) digging and filling the bucket, (2) transporting materials, and (3) unloading. A natural approach is to employ object detection models such as YOLO \cite{redmon2016you}, which have been widely adopted in areas like autonomous driving \cite{sarda2021object} and disaster management \cite{prabhu2022rescuenet}. However, as illustrated in Figure \ref{fig:badex}, such models frequently suffer from robustness issues when faced with conditions not represented in the training set. Since collecting exhaustive training data is impractical, prediction errors are often inevitable. Common workarounds—such as raising confidence thresholds—introduce a trade-off between missing detections and accepting false positives.

Initially, the leasing company implemented an excavator monitoring system using a YOLO model to detect buckets and trucks, complemented by manually designed heuristic rules to identify complete work cycles. These rules essentially set thresholds for counting occurrences of vertically and horizontally oriented buckets. However, regardless of how carefully these rules were tuned, whether strict or lenient, failures persisted due to the inherent inflexibility of applying fixed rules to highly variable real-world scenarios.

To resolve this dilemma, we introduce an FSM-driven streaming inference pipeline that robustly encapsulates the operational workflow, replacing the brittle manual rules. Finite State Machines have been widely adopted across diverse domains, demonstrating their versatility in modeling complex processes. For instance, Salem et al. \cite{salem2016practical} developed an FSM-based library for industrial programming, validation, and verification. Jagdale \cite{jagdale2021finite} applied FSMs to manage game states and behaviors. And Wang et al. \cite{wang2017modeling} modeled hot stamping processes using FSMs, underscoring their applicability in manufacturing. Inspired by these successes, we leverage an FSM to bridge business logic with object detection, thereby enhancing the reliability of machine learning applications in dynamic environments.
Experimental results show that our approach achieves an approximately 2\% higher F1-score than the original solution while delivering substantially more stable performance, with error rates consistently bounded within a narrow margin.

\section{Approach}

\begin{figure*}[htbp]
\centerline{\includegraphics[width=\textwidth]{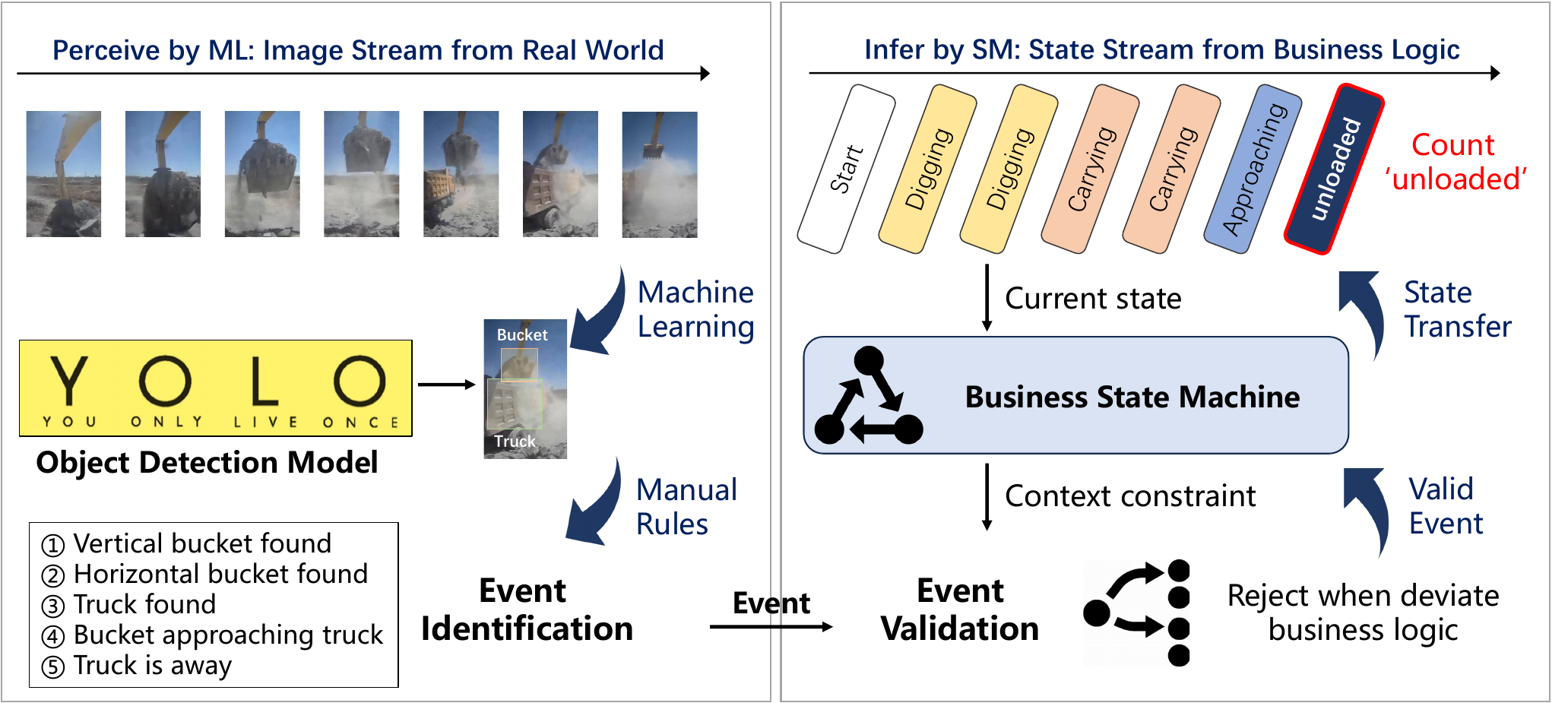}}
\caption{The overview of our approach}
\label{method}
\end{figure*}

The proposed approach comprises two main parts, as shown in Figure \ref{method}.
The first part identifies events using object detection techniques. 
We employ a machine learning model, i.e., YOLOv9 \cite{yolov9}, that repeatedly processes real-world data sequences and identifies the occurring events.
The second part uses a finite state machine to infer the next business state from the current state and to filter out invalid or noisy events produced by the detector.

\subsection{Event Identification by Object Detection}
Our approach is based on object detection, a standard machine learning task that locates and recognizes target objects in images.
The output of object detection models includes the tags and locations of target objects, enabling us to determine which object is in the image and where it is located. 
In the application of monitoring excavators, our target objects include buckets and trucks.
In particular, buckets are classified as vertical or horizontal to fully describe what is happening in the image.

We use \textit{event} as the signals that indicate changes in the working process of excavators.
For example, the emergence of a specific object or the formation of a specific positional relation between two objects can both be events.
We extract useful information about target objects from images using state-of-the-art object detection models, such as YOLOv9, to identify events.

We first define three simple events, such as \textit{vertical bucket found ($e_0$)}, \textit{horizontal bucket found ($e_1$)}, and \textit{truck found ($e_2$)}.
When a vertical bucket is detected in an image, event $e_0$ is triggered. The same is true for events $e_1$ and $e_2$.
We also define two complex events that exploit the positional relationships between trucks and buckets, namely \textit{Bucket approaching truck ($e_3$)} and \textit{Truck is away ($e_4$)}.
The events are presented as follows:
\begin{itemize}
    \item \textbf{Vertical bucket found ($e_0$).} Once vertical buckets are detected in the image.
    \item \textbf{Horizontal bucket found ($e_1$).} Once horizontal buckets are detected in the image.
    \item \textbf{Truck found ($e_2$).} Once trucks are detected in the image.
    \item \textbf{Bucket approaching truck ($e_3$).} Trucks and buckets are both detected in the same images more than twice, and the distance between the bucket and the truck is getting smaller. Notably, when more than one bucket or truck is detected, the biggest one in the image will be used to calculate the distance.
    \item \textbf{Truck is away ($e_4$).} Trucks and buckets emerge in the same images more than twice, and the distance between the bucket and the truck is getting bigger. Or after emerging once, the truck is missing from the adjacent images. 
\end{itemize}

We consider identifying events from original images to be an end-to-end process consisting of the object detection model and the rules for using object detection results to define events.
The rules can be designed carefully to be accurate and robust enough. However, uncertainty from the object detection models is difficult to avoid manually.
A method to prevent failures caused by wrongly identified events is presented in the next section.

\subsection{State Transfer By Business State Machine}

In this section, we use a finite state machine to construct business logic, which utilizes the above-identified events to drive the transfer of business state.

\subsubsection{Business States}
The state machine method is widely used because it can simplify complex processes into separate states and bridge them with specific events.
We use a state machine to divide one workload of excavators into 5 business states to cover commonly seen working situations of excavators.

We first summarise the excavators' working process.
Generally, one workload starts with digging the ground, and the bucket is thereafter filled with material like dirt.
The full bucket is then lifted and moved to a truck for transport.

A first intuition design of one workload's business states is {Digging ($s_0$)}, {Carrying ($s_1$)}, and {Unloaded ($s_4$)}.
However, \textit{unload} can not be easily identified through object detection.
Limited by the camera conditions, the unloading movement is often out of the camera's vision.
To address the problem, we supplement another state \textit{Approaching ($s_2$)}, since the movement of moving buckets to trucks can be easily captured.
But it is not the end.
In many situations, it is not certain that one workload is definitely finished after the state \textit{Approaching ($s_2$)}, so we supplement another state \textit{Possibly Unloaded ($s_3$)}.
Therefore, we propose 5 business states to describe the working of excavators as follows:
\begin{itemize}
    \item \textbf{Digging ($s_0$).} The excavator is digging dirt.
    \item \textbf{Carrying ($s_1$).} The bucket is filled with dirt and is carrying dirt somewhere. 
    \item \textbf{Approaching ($s_2$).} The bucket is moving towards a truck.
    \item \textbf{Possibly unloaded ($s_3$).} The dirt in the bucket may be unloaded into the truck.
    \item \textbf{Unloaded ($s_4$).} The dirt in the bucket is sure to be unloaded into the truck.
\end{itemize}

\begin{figure}[!t]
\centering
\includegraphics[width=\linewidth,height=0.28\textheight,keepaspectratio]{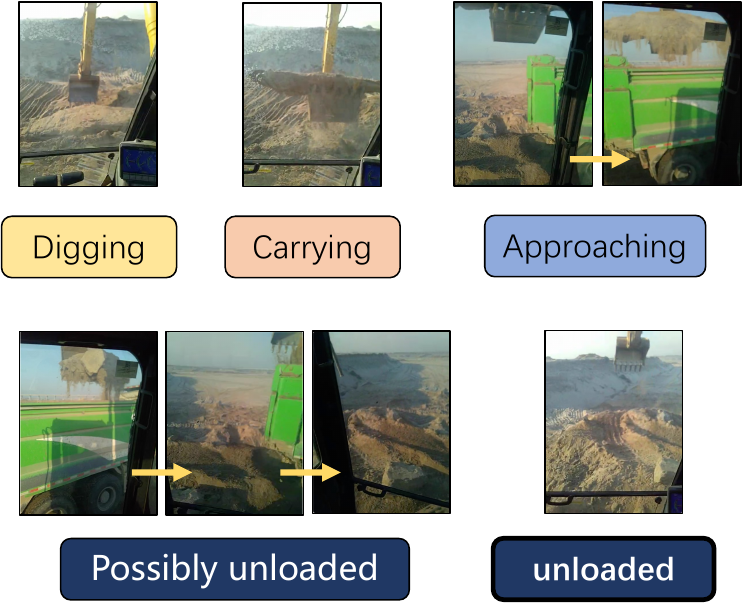}
\caption{The business states of excavator monitoring}
\label{states}
\end{figure}

The typical images or image sequences of the states are as presented in Figure \ref{states}.
There can also be other ways to define states, and we adopt this in this paper as an example.

\subsubsection{State Transfer and Event Rectification}
In our approach, the detector perceives information from the real world, fully exploiting the advantages of machine learning technology.
Machine learning still encounters difficulties when dealing with complex processes.
Thus, we introduce the state machine to divide the complex process into several states, breaking down the demanding job into a sequence of more straightforward tasks.
The state machine is constructed based on the events identified by the object detection model.
We denote the state machine as a function $f$, then $s_{t+1} = f(s_t, e_t)$, where $t$ means the current timestamp and $t+1$ means the timestamp after event $e_t$ happened.

\begin{figure}[!t]
\centering
\includegraphics[width=\linewidth,height=0.28\textheight,keepaspectratio]{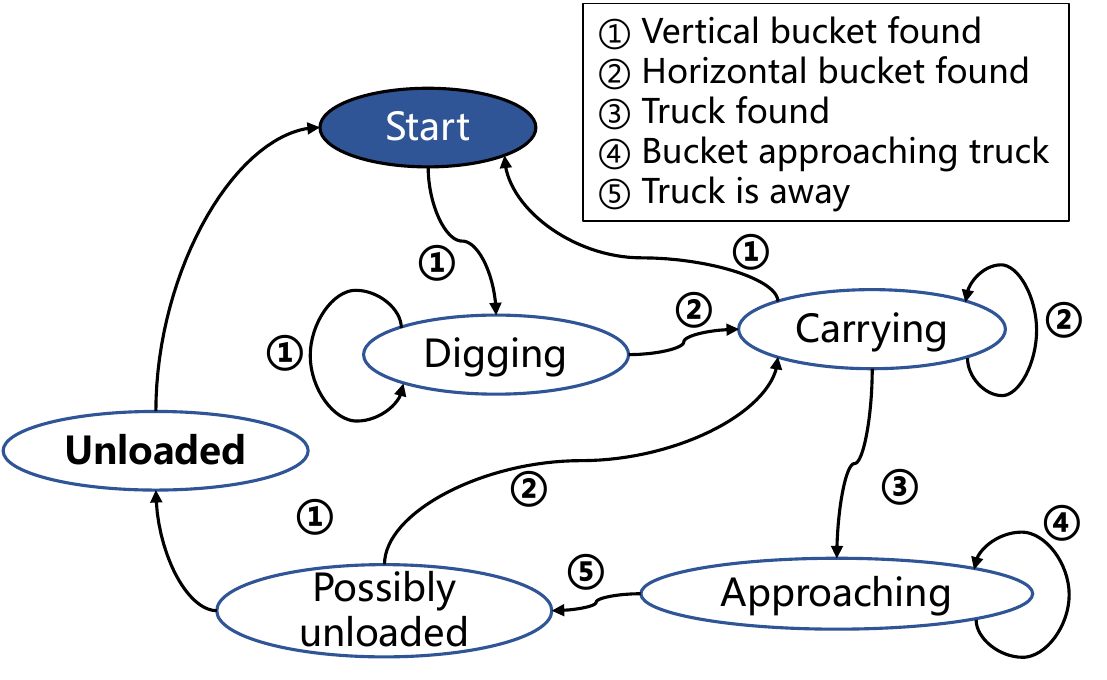}
\caption{The business states of excavator monitoring}
\label{machine}
\end{figure}

The development of the state machine is non-trivial.
First, it should cover the entire process and main branches of business, which must be thoroughly debugged using various real-world data samples.
Second, an unambiguous bound among the states is necessary to identify recognizable events supported by machine learning tasks.
The greater the gap between two adjacent states, the more likely a machine learning model is to identify the state transfer event.
Third, the results of the machine learning models should intuitively identify the state transfer event without relying on other ambiguous information, as uncertainty may diminish their effectiveness.
The whole business logic presented by the state machine is shown in Figure \ref{machine}.

In the original solutions, the inference function $g$ is based on manual heuristic rules, and $g(\{e_t,e_{t-1},...e_{t-k})$ returns whether a complete workload finished, where $e_{t-k}$ is the beginning of the workload.
The working process of excavators is seemingly monotonous and straightforward, but many unexpected situations can occur in the wild, potentially leading to incorrect predictions by machine learning models.
When the wrong predictions happen, the event $e_t$ is possibly incorrect as a result.
Thus, $g(\{e_t,e_{t-1},...e_{t-k})$ has a high probability of being influenced by wrong predictions because any event in $\{e_t,e_{t-1},...e_{t-k}$ may fail $g(\{e_t,e_{t-1},...e_{t-k})$.

Instead, we utilize the state machine to provide a valid event space to prevent failures caused by wrongly identified events.
For example, in the state of $s_{Digging}$, only events $e_{horizontal}$ and $e_{vertical}$ are valid events, and the state machine will ignore other events.
Every time the detector identifies an event, the state machine checks whether the state transition should be triggered.
In this way, $s_{t+1} = f(s_t, e_t)$ is much robust than $g(\{e_t,e_{t-1},...e_{t-k})$.
When the state $s_{unloaded}$ is achieved, one complete workload is supposed to be finished, and the state returns to the start.

\subsection{Data Engineering}

This subsection details the data engineering workflow that supports our excavator workload monitoring system. Our objective is not only to obtain good offline accuracy, but also to keep the system stable under diverse, noisy, and dynamic site conditions. We therefore describe dataset construction, annotation protocol, preprocessing, split policy, and key engineering trade-offs.

\subsubsection{Dataset Construction.}
We collect videos from multiple construction sites, covering diverse working conditions such as strong sunlight, backlight, dust, rain, and nighttime operation. Camera installations differ in height (4--12\,m), view angle (15--40$^\circ$), and focal length, which naturally introduces variations in scale, occlusion, and visual clutter.  
The resulting dataset contains approximately 16,000 training images and 7,000 evaluation images extracted from 24–30\, FPS videos. To reduce redundancy while maintaining motion coverage, frames are sampled with stride $s\in\{3,5,10\}$. This ensures that short-term motion changes—such as bucket lifting or truck approach—are preserved.  
Three object classes are defined: \textit{Truck}, \textit{Bucket-Vertical}, and \textit{Bucket-Horizontal}. Instead of identifying individual buckets, we focus on pose states, because the FSM logic depends on pose transitions to determine valid workloads rather than object identity.

\subsubsection{Annotation and Quality Control}
Each sampled frame is independently labeled by two annotators and checked by a third when disagreements occur. Annotators are instructed to draw tight bounding boxes and decide the bucket pose based on the lip orientation relative to the ground. Regular quality audits are performed to maintain consistency across sessions.  
A lightweight temporal validator automatically detects improbable pose sequences: if a bucket alternates between vertical and horizontal states within 0.2\,s, the sequence is flagged for review. This catches transient label noise caused by blur, motion, or partial occlusion.  
Furthermore, we adopt a continuous data refinement loop: logs from real deployments are analyzed to identify systematic failure cases—such as missed detections under backlight or confusing clutter—and corresponding clips are re-labeled and appended to the training data. This process gradually improves both dataset coverage and model reliability.

\subsubsection{Preprocessing Pipeline}
Raw videos are decoded into RGB frames, resized to $640\times640$ using letterbox padding to maintain aspect ratio, and normalized to $[0,1]$.  
We apply light data augmentations, including brightness and contrast jitter ($\pm10\%$), mild Gaussian blur to mimic dusty conditions, and small affine transformations ($<5^\circ$ rotation, $<10\%$ scale). Heavy geometric warping is avoided to prevent altering the bucket’s true pose semantics.  
To reduce redundant data, near-duplicate frames are removed based on SSIM$>0.98$ within a one-second window. This step lowers labeling cost without losing temporal continuity. All preprocessed images, annotations, and split lists are version-controlled to guarantee reproducibility.

\subsubsection{Data Splitting and Leakage Prevention}
To avoid information leakage between training and evaluation, we partition data by site and recording date. Frames or clips from the same site-day are never distributed across multiple splits. For long continuous videos, we adopt block-wise partitioning (e.g., 60\%/20\%/20\% for train/val/test) instead of random frame selection.  
Each workload—defined as a complete truck loading cycle—is treated as an atomic unit, wholly contained in a single split. This prevents partial workloads from leaking into both train and test sets, which could otherwise inflate performance metrics.  
Such careful segmentation ensures that evaluation reflects generalization to unseen environments rather than memorization of visual patterns.

\subsubsection{Engineering Trade-offs}
To achieve a practical balance between accuracy, latency, and robustness, several engineering choices are tuned at deployment time. The following design factors summarize the main trade-offs considered during system development and field adaptation.

\begin{itemize}
    \item \textbf{Threshold tuning.} The confidence $\tau_c$ and IoU $\tau_{IoU}$ are optimized on the validation set. Higher $\tau_c$ reduces false positives in cluttered or bright scenes but may reduce recall under low visibility. Site-specific tuning provides balanced precision and recall, allowing the detector to adapt to differing illumination and dust levels across sites.

    \item \textbf{Temporal stride.} Increasing the stride improves throughput but risks skipping short-duration actions. In practice, $k{=}5$ achieves real-time processing with negligible accuracy loss, because the FSM integrates evidence across multiple frames. This choice balances computational efficiency with temporal fidelity for dynamic workloads.

    \item \textbf{One-per-class policy.} Retaining only the top-1 detection for each class simplifies downstream reasoning and avoids contradictory transitions when multiple machines coexist. It also improves FSM stability by ensuring deterministic state updates even under overlapping detections.

    \item \textbf{Logging and reproducibility.} During deployment, each processed frame logs the timestamp, FSM state, detected event, and confidence scores. Together with fixed random seeds and version-controlled manifests, this ensures full traceability for debugging, retraining, and benchmarking. Such logging also supports automated error discovery and continuous data improvement.
\end{itemize}

The proposed data engineering framework integrates systematic sampling, high-quality annotation, controlled augmentation, careful split management, and FSM-aware deployment settings to provide a robust basis for reliable workload detection. These practices keep behavior consistent across diverse environments, enable continuous data improvement, and support reproducible benchmarking for future deployments.

\section{EVALUATION}
In this section, we present the evaluation of the proposed method.
Our solution is composed of an object detection model and a state machine.

\textbf{RQ1. What is the performance of the proposed method?} 
We use 12 videos (more than 7,000 images) of working excavators collected from real construction sites to evaluate the performance of the proposed methods.
The motivation of the application is to count workloads while working with an excavator.
However, a close number does not definitely mean that the solution is good.
Wrong evaluations can result from the neutralization of missing workloads and fake workloads, so both precision and recall are considered in the assessment.

\textbf{RQ2. What is the reliability of the proposed method?} 
Before our approach, the solution based on heuristic rules had achieved a competitive performance.
However, the manual heuristic rules often lead to a few unacceptable bad cases. 
We further investigate some experiments to show that our approach achieves much more stable performance than manual heuristic rules. 
\subsection{RQ1. Overall Performance}

\subsubsection{Settings}

We train the object detection model with about 16,000 images that are sampled from real construction sites.
The images are labeled with vertical buckets, horizontal buckets, and trucks, together with their positions.
And we use another 7,000 images from 12 videos in real construction sites to test the solutions.

As a baseline method, we use the solution based on heuristic rules instead of the state machine to count the number of workloads an excavator has made.
Both the baseline and our approach utilize the object detection results to identify events.
The baseline records the number of vertical and horizontal bucket events. When the cumulations reach thresholds, one pass of the working process is counted. 
In particular, we reset the vertical bucket event counting whenever a horizontal bucket event is identified. 
This setting came from practical experience, showing that vertical buckets emerge more frequently than horizontal buckets. Therefore, horizontal buckets play a key role in recognizing each complete loading process.

We use both precision, recall, and F1-score as metrics in evaluation.
We use \textit{Tr} to denote the count of ground truth, \textit{CT} to denote the count of the baseline and our approach, \textit{P} to denote precision, \textit{R} to denote recall, and \textit{F1} to denote F1-score.

\subsubsection{Results}

The results are presented in Table \ref{effectiveness}. 
Notably, the $CT$ of the baseline and our approach are both very close to $Tr$.
However, this does not mean the counting is accurate because when a fake workload and a missed workload occur together, $|CT-Tr|$ will not be changed.
From the precision and recall, we can see that some mistakes are neutralized.
This is the reason we compare the methods using the F1-score instead of just the counting results.
The heuristic rules are well-tuned throughout the videos to achieve a strong baseline.
Even so, our approach performs better with much higher precision and comparable recall.
The advantage of our approach on the F1-score is 2 percent.

\begin{table}[htbp]
\caption{Comparison of excavation operation counts using heuristic rules and state machine.}
\label{effectiveness}
\centering
\footnotesize
\begin{tabular}{@{}cc|cccc|cccc@{}}
\toprule
\multicolumn{2}{c|}{\textbf{Test Video}} &
\multicolumn{4}{c|}{\textbf{Heuristic Rules}} &
\multicolumn{4}{c}{\textbf{State Machine}} \\
\cmidrule(lr){1-2}\cmidrule(lr){3-6}\cmidrule(lr){7-10}
\textbf{NO.} & \textbf{Tr} & \textbf{CT} & \textbf{P} & \textbf{R} & \textbf{F1}
            & \textbf{CT} & \textbf{P} & \textbf{R} & \textbf{F1} \\
\midrule
1  & 39 & 40 & 0.95 & \textbf{0.97} & 0.96 & 38 & \textbf{1.00} & \textbf{0.97} & \textbf{0.99}\\
2  & 35 & 35 & \textbf{1.00} & \textbf{1.00} & \textbf{1.00} & 34 & \textbf{1.00} & 0.97 & 0.99 \\
3  & 32 & 39 & 0.82 & \textbf{1.00} & 0.90 & 36 & \textbf{0.89} & \textbf{1.00} & \textbf{0.94} \\
4  & 32 & 34 & 0.94 & \textbf{1.00} & \textbf{0.97} & 32 & \textbf{0.97} & 0.97 & \textbf{0.97} \\
5  & 31 & 35 & 0.89 & \textbf{1.00} & 0.94 & 31 & \textbf{1.00} & \textbf{1.00} & \textbf{1.00} \\
6  & 30 & 31 & \textbf{0.97} & \textbf{1.00} & \textbf{0.98} & 31 & \textbf{0.97} & \textbf{1.00} & \textbf{0.98} \\
7  & 29 & 24 & \textbf{1.00} & 0.83 & \textbf{0.91} & 26 & 0.96 & \textbf{0.86} & \textbf{0.91} \\
8  & 27 & 31 & 0.84 & \textbf{0.96} & 0.90 & 26 & \textbf{0.96} & \textbf{0.93} & \textbf{0.94} \\
9  & 25 & 20 & 0.95 & 0.76 & 0.84 & 20 & \textbf{1.00} & \textbf{0.80} & \textbf{0.89} \\
10 & 24 & 27 & 0.91 & \textbf{0.96} & \textbf{0.90} & 21 & \textbf{0.95} & 0.83 & 0.89 \\
11 & 23 & 21 & 0.95 & 0.86 & 0.91 & 20 & \textbf{1.00} & \textbf{0.87} & \textbf{0.93} \\
12 & 22 & 28 & 0.75 & \textbf{0.95} & 0.84 & 24 & \textbf{0.83} & 0.91 & \textbf{0.87} \\
\midrule
AVG & 29.1 & 30.4 & 0.91 & \underline{\textbf{0.94}} & 0.92
    & 28.3 & \underline{\textbf{0.96}} & 0.93 & \underline{\textbf{0.94}} \\
\bottomrule
\end{tabular}
\end{table}

\noindent\textbf{Answering RQ1:}
The state machine outperforms the carefully tuned manual heuristic rules when used as business logic.
Specifically, precision improved from 0.91 to 0.96, recall degraded slightly from 0.94 to 0.93, and the F1 score increased from 0.92 to 0.94.

\subsection{RQ2. Advantages of Stability}

\subsubsection{Settings}
In this section, we follow the above settings of RQ1.
The results of RQ1 show that heuristic rules perform poorly on precision but well on recall.
It is partially because the heuristic rules are relatively loose, and many wrongly identified events lead to the counting of fake workloads.
We design another baseline with strict heuristic rules to relieve this issue.
Then, we compare the counted fake workloads and the missing workloads among the state machine, the strict rules, and the loose rules to show the stability of our approach.

\subsubsection{Results}

\begin{figure}[!tb]
\centering
\begin{minipage}[t]{0.48\linewidth}
  \centering
  \includegraphics[width=\linewidth,height=0.18\textheight,keepaspectratio]{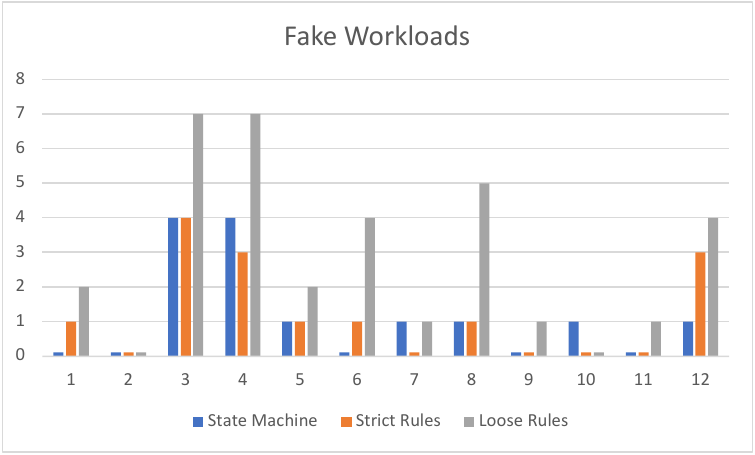}\\
  \vspace{2pt}
  (a) Fake workloads
\end{minipage}\hfill
\begin{minipage}[t]{0.48\linewidth}
  \centering
  \includegraphics[width=\linewidth,height=0.18\textheight,keepaspectratio]{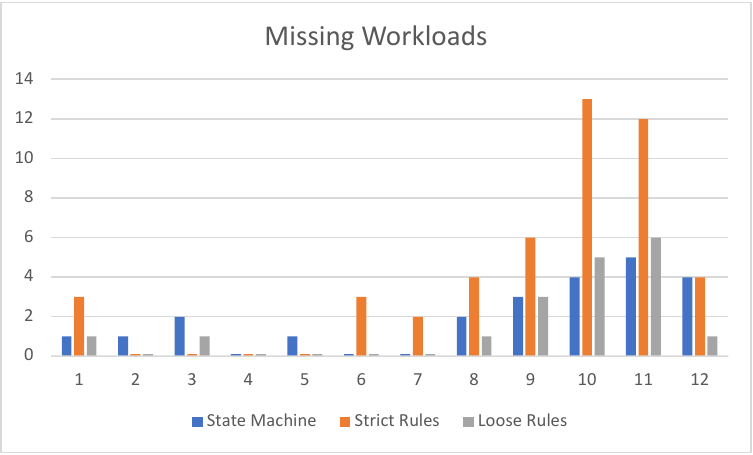}\\
  \vspace{2pt}
  (b) Missing workloads
\end{minipage}
\caption{Comparison of mistakes under different business logic.}
\label{fig:rq2-mistakes}
\end{figure}

The results are presented in Figure~\ref{fig:rq2-mistakes}(a) andFigure~\ref{fig:rq2-mistakes}(b).
The strict rules perform well on fake workloads, achieving a precision of 0.958, which is comparable to our approach's 0.961.
However, for missing workloads, the recall of the strict rules is only 0.855, which is much smaller than our approach's 0.926.
The loose rules perform well on missing workloads, achieving a recall of 0.942, which is slightly better than our approach's 0.926.
However, for fake workloads, the precision of the loose rules is only 0.909, which is much smaller than the 0.961 of our approach.

We can see that, compared with the state machine, the carefully tuned heuristic rules can overfit some exceptional cases to achieve comparable performance.
However, such overfitting always leads to weakness on the other side of the target.
Just like the training of machine learning models, overfitting always indicates bad generalizability.

Instead, the state machine performs consistently well.
Our approach prevents overfitting by using the state machine, which decomposes the complex working process into a series of separated business states.
The inference for each state is much simpler and more stable than the global inference based on the manual heuristic rules.

\noindent\textbf{Answering RQ2:}
The state machine is more reliable than the carefully tuned manual heuristic rules, whether strict or loose.
Specifically, the number of counted fake workloads is 13, fewer than the 14 under the strict rules and 34 under the loose regulations.
The number of missing workloads is 23, which is much less than the 47 under the strict rules but five more than the 18 under the loose regulations.
On the whole, our approach ranks the best, considering both counted fake workloads and missing workloads.

\section{Related Work}
In recent years, multimodal large language models (MLLMs), represented by GPT-4V, have become an emerging research hotspot and are widely applied in various fields such as vision-language understanding\cite{awadalla2023openflamingo}, video processing\cite{li2024mvbench}, biomedicine\cite{li2024llava}, and document parsing\cite{ye2023mplug}.
With the rise of the large multimodal language model, an end-to-end video-based solution has become possible.

We have attempted to use large multimodal models\cite{li2024mvbench} to analyze videos and calculate the number of workloads an excavator has done.
To simplify the problem, we put the video into several small segments.
Then we have a dialogue with the multimodal large model to tell the background and our motivation.

However, the results were unsatisfactory, even for some obvious situations. 
This could be due to the extended time required for the excavator to complete a single workload.
The model's limited capacity constrains the processing to only a small number of frames at once, leading to a lower frame sampling rate.
Thus, information is easily lost.
Besides, we did not fine-tune the model since the data is insufficient.
The model's ability to recognize buckets and trucks may still be poor.
Totally, an end-to-end solution to such problems is still hard to achieve.

Several works consider utilizing knowledge like rules and physical theories when training machine learning models \cite{karniadakis_physics-informed_2021}.
Abbasi et al. \cite{abbasi_physics-informed_2022} measure the confidence level of the data-driven model on its predictions and use the physics-based model proportional to the uncertainty of those predictions.
Lin et al. \cite{lin_hybrid_2021} build a hybrid physics-based and data-driven modeling for online diagnosis and prognosis of battery degradation, achieving better generalizability and interpretability together with a well-calibrated uncertainty associated with its prediction.
Such methods are also called physics-informed neural networks (PINN) \cite{huang_applications_2022}.

\section{CONCLUSION}

In this paper, we propose an approach to reliably present business logic using object detection models and finite state machines.
Compared with the original solution based on manual heuristic rules, our approach prevents overfitting to some exceptional cases and reduces the risk of failures caused by wrong object detection predictions.
We conduct experiments on 12 videos from real construction sites to evaluate the effectiveness of the approach. 
The results show that our approach performs much better and is more robust. 

\bibliographystyle{iclr2026_conference}
\bibliography{iclr2026_conference}

@String{Computing = "Computing" }

@String{Computer = "{IEEE} Computer" }

@String{Springer = "Springer-Verlag" }

@ArtifactSoftware{R,
    title = {R: A Language and Environment for Statistical Computing},
    author = {{R Core Team}},
    organization = {R Foundation for Statistical Computing},
    address = {Vienna, Austria},
    year = {2019},
    url = {https://www.R-project.org/},
}

@inproceedings{nfr_2021,
	title = {Non-functional {Requirements} for {Machine} {Learning}: {Understanding} {Current} {Use} and {Challenges} in {Industry}},
	doi = {10.1109/RE51729.2021.00009},
	booktitle = {2021 {IEEE} 29th {International} {Requirements} {Engineering} {Conference} ({RE})},
	author = {Habibullah, Khan Mohammad and Horkoff, Jennifer},
	month = sep,
	year = {2021},
	note = {ISSN: 2332-6441},
	pages = {13--23},
}

@article{huang_applications_2022,
	title = {Applications of {Physics}-{Informed} {Neural} {Networks} in {Power} {Systems} - {A} {Review}},
	issn = {1558-0679},
	doi = {10.1109/TPWRS.2022.3162473},
	journal = {IEEE Transactions on Power Systems},
	author = {Huang, Bin and Wang, Jianhui},
	year = {2022},
	note = {Conference Name: IEEE Transactions on Power Systems},
	pages = {1--1},
}

@misc{lin_hybrid_2021,
	title = {Hybrid physics-based and data-driven modeling with calibrated uncertainty for lithium-ion battery degradation diagnosis and prognosis},
	url = {http://arxiv.org/abs/2110.13661},
	doi = {10.48550/arXiv.2110.13661},
	urldate = {2022-09-03},
	publisher = {arXiv},
	author = {Lin, Jing and Zhang, Yu and Khoo, Edwin},
	month = nov,
	year = {2021},
	note = {arXiv:2110.13661 [physics]},
}

@article{karniadakis_physics-informed_2021,
	title = {Physics-informed machine learning},
	volume = {3},
	copyright = {2021 Springer Nature Limited},
	issn = {2522-5820},
	url = {https://www.nature.com/articles/s42254-021-00314-5},
	doi = {10.1038/s42254-021-00314-5},
	language = {en},
	number = {6},
	urldate = {2022-05-16},
	journal = {Nature Reviews Physics},
	author = {Karniadakis, George Em and Kevrekidis, Ioannis G. and Lu, Lu and Perdikaris, Paris and Wang, Sifan and Yang, Liu},
	month = jun,
	year = {2021},
	note = {Number: 6
Publisher: Nature Publishing Group},
	pages = {422--440},
}

@article{abbasi_physics-informed_2022,
	title = {Physics-{Informed} {Machine} {Learning} for {Uncertainty} {Reduction} in {Time} {Response} {Reconstruction} of a {Dynamic} {System}},
	volume = {26},
	issn = {1941-0131},
	doi = {10.1109/MIC.2022.3170736},
	number = {4},
	journal = {IEEE Internet Computing},
	author = {Abbasi, Amirhassan and Nataraj, C.},
	month = jul,
	year = {2022},
	note = {Conference Name: IEEE Internet Computing},
	pages = {35--44},
}

@misc{yolov9,
      title={YOLOv9: Learning What You Want to Learn Using Programmable Gradient Information}, 
      author={Chien-Yao Wang and I-Hau Yeh and Hong-Yuan Mark Liao},
      year={2024},
      eprint={2402.13616},
      archivePrefix={arXiv},
      primaryClass={cs.CV},
      url={https://arxiv.org/abs/2402.13616}, 
}

@article{app,
    title = {Machine Learning for industrial applications: A comprehensive literature review},
    journal = {Expert Systems with Applications},
    volume = {175},
    pages = {114820},
    year = {2021},
    issn = {0957-4174},
    doi = {https://doi.org/10.1016/j.eswa.2021.114820},
    url = {https://www.sciencedirect.com/science/article/pii/S095741742100261X},
    author = {Massimo Bertolini and Davide Mezzogori and Mattia Neroni and Francesco Zammori},
}

@misc{reg,
      title={Regularization for Deep Learning: A Taxonomy}, 
      author={Jan Kukačka and Vladimir Golkov and Daniel Cremers},
      year={2017},
      eprint={1710.10686},
      archivePrefix={arXiv},
      primaryClass={cs.LG},
      url={https://arxiv.org/abs/1710.10686}, 
}

@misc{at,
      title={Recent Advances in Adversarial Training for Adversarial Robustness}, 
      author={Tao Bai and Jinqi Luo and Jun Zhao and Bihan Wen and Qian Wang},
      year={2021},
      eprint={2102.01356},
      archivePrefix={arXiv},
      primaryClass={cs.LG},
      url={https://arxiv.org/abs/2102.01356}, 
}

@misc{da,
      title={The Effectiveness of Data Augmentation in Image Classification using Deep Learning}, 
      author={Luis Perez and Jason Wang},
      year={2017},
      eprint={1712.04621},
      archivePrefix={arXiv},
      primaryClass={cs.CV},
      url={https://arxiv.org/abs/1712.04621}, 
}

@article{qa_2022,
	title = {Construction of a quality model for machine learning systems},
	volume = {30},
	issn = {1573-1367},
	url = {https://doi.org/10.1007/s11219-021-09557-y},
	doi = {10.1007/s11219-021-09557-y},
	language = {en},
	number = {2},
	urldate = {2023-02-27},
	journal = {Software Quality Journal},
	author = {Siebert, Julien and Joeckel, Lisa and Heidrich, Jens and Trendowicz, Adam and Nakamichi, Koji and Ohashi, Kyoko and Namba, Isao and Yamamoto, Rieko and Aoyama, Mikio},
	month = jun,
	year = {2022},
	pages = {307--335},
}

@inproceedings{qa_2020,
	title = {Requirements-{Driven} {Method} to {Determine} {Quality} {Characteristics} and {Measurements} for {Machine} {Learning} {Software} and {Its} {Evaluation}},
	doi = {10.1109/RE48521.2020.00036},
	booktitle = {2020 {IEEE} 28th {International} {Requirements} {Engineering} {Conference} ({RE})},
	author = {Nakamichi, Koji and Ohashi, Kyoko and Namba, Isao and Yamamoto, Rieko and Aoyama, Mikio and Joeckel, Lisa and Siebert, Julien and Heidrich, Jens},
	year = {2020},
	note = {ISSN: 2332-6441},
	pages = {260--270},
}

@misc{mlops2022,
      title={Machine Learning Operations (MLOps): Overview, Definition, and Architecture}, 
      author={Dominik Kreuzberger and Niklas Kühl and Sebastian Hirschl},
      year={2022},
      eprint={2205.02302},
      archivePrefix={arXiv},
      primaryClass={cs.LG},
      url={https://arxiv.org/abs/2205.02302}, 
}

@inproceedings{PeiLWW22,
  author       = {Zhongyi Pei and
                  Lin Liu and
                  Chen Wang and
                  Jianmin Wang},
  title        = {Requirements Engineering for Machine Learning: {A} Review and Reflection},
  booktitle    = {30th {IEEE} International Requirements Engineering Conference Workshops,
                  {RE} 2022 - Workshops, Melbourne, Australia, August 15-19, 2022},
  pages        = {166--175},
  publisher    = {{IEEE}},
  year         = {2022},
  url          = {https://doi.org/10.1109/REW56159.2022.00039},
  doi          = {10.1109/REW56159.2022.00039},
  bibsource    = {dblp computer science bibliography, https://dblp.org}
}

@inproceedings{salem2016practical,
  title={Practical programming, validation and verification with finite-state machines: a library and its industrial application},
  author={Salem, Paulo},
  booktitle={Proceedings of the 38th International Conference on Software Engineering Companion},
  pages={51--60},
  year={2016}
}

@article{jagdale2021finite,
  title={Finite state machine in game development},
  author={Jagdale, Devang},
  journal={International Journal of},
  year={2021}
}

@article{wang2017modeling,
  title={Modeling of hot stamping process procedure based on finite state machine (FSM)},
  author={Wang, Liang and Zhu, Bin and Wang, Qiang and Zhang, Yisheng},
  journal={The International Journal of Advanced Manufacturing Technology},
  volume={89},
  pages={857--868},
  year={2017},
  publisher={Springer}
}

@inproceedings{redmon2016you,
  title={You only look once: Unified, real-time object detection},
  author={Redmon, Joseph and Divvala, Santosh and Girshick, Ross and Farhadi, Ali},
  booktitle={Proceedings of the IEEE conference on computer vision and pattern recognition},
  pages={779--788},
  year={2016}
}

@inproceedings{sarda2021object,
  title={Object detection for autonomous driving using yolo [you only look once] algorithm},
  author={Sarda, Abhishek and Dixit, Shubhra and Bhan, Anupama},
  booktitle={2021 Third international conference on intelligent communication technologies and virtual mobile networks (ICICV)},
  pages={1370--1374},
  year={2021},
  organization={IEEE}
}

@article{prabhu2022rescuenet,
  title={RescueNet: YOLO-based object detection model for detection and counting of flood survivors},
  author={Prabhu, BV Balaji and Lakshmi, R and Ankitha, R and Prateeksha, MS and Priya, NC},
  journal={Modeling Earth Systems and Environment},
  volume={8},
  number={4},
  pages={4509--4516},
  year={2022},
  publisher={Springer}
}

@article{awadalla2023openflamingo,
  title={Openflamingo: An open-source framework for training large autoregressive vision-language models},
  author={Awadalla, Anas and Gao, Irena and Gardner, Josh and Hessel, Jack and Hanafy, Yusuf and Zhu, Wanrong and Marathe, Kalyani and Bitton, Yonatan and Gadre, Samir and Sagawa, Shiori and others},
  journal={arXiv preprint arXiv:2308.01390},
  year={2023}
}

@article{li2024llava,
  title={Llava-med: Training a large language-and-vision assistant for biomedicine in one day},
  author={Li, Chunyuan and Wong, Cliff and Zhang, Sheng and Usuyama, Naoto and Liu, Haotian and Yang, Jianwei and Naumann, Tristan and Poon, Hoifung and Gao, Jianfeng},
  journal={Advances in Neural Information Processing Systems},
  volume={36},
  year={2024}
}

@article{ye2023mplug,
  title={mplug-docowl: Modularized multimodal large language model for document understanding},
  author={Ye, Jiabo and Hu, Anwen and Xu, Haiyang and Ye, Qinghao and Yan, Ming and Dan, Yuhao and Zhao, Chenlin and Xu, Guohai and Li, Chenliang and Tian, Junfeng and others},
  journal={arXiv preprint arXiv:2307.02499},
  year={2023}
}

@inproceedings{li2024mvbench,
  title={Mvbench: A comprehensive multi-modal video understanding benchmark},
  author={Li, Kunchang and Wang, Yali and He, Yinan and Li, Yizhuo and Wang, Yi and Liu, Yi and Wang, Zun and Xu, Jilan and Chen, Guo and Luo, Ping and others},
  booktitle={Proceedings of the IEEE/CVF Conference on Computer Vision and Pattern Recognition},
  pages={22195--22206},
  year={2024}
}

@article{canhoto2020artificial,
  title={Artificial intelligence and machine learning as business tools: A framework for diagnosing value destruction potential},
  author={Canhoto, Ana Isabel and Clear, Fintan},
  journal={Business Horizons},
  volume={63},
  number={2},
  pages={183--193},
  year={2020},
  publisher={Elsevier}
}

@article{nalchigar2018business,
  title={Business-driven data analytics: A conceptual modeling framework},
  author={Nalchigar, Soroosh and Yu, Eric},
  journal={Data \& Knowledge Engineering},
  volume={117},
  pages={359--372},
  year={2018},
  publisher={Elsevier}
}

\end{document}